\pdfoutput=1

\documentclass[11pt]{article}
\usepackage[]{acl}

\usepackage{times}
\usepackage{latexsym}

\usepackage{array}
\usepackage{tabularx}
\usepackage{hyperref}

\usepackage{listings}
\usepackage[T1]{fontenc}
\usepackage{multirow}

\usepackage[utf8]{inputenc}

\usepackage{caption}
\usepackage{subcaption}

\usepackage{microtype}

\usepackage{inconsolata}

\usepackage{graphicx}
\usepackage{pgfplots}
\pgfplotsset{compat=1.17}
\title{Can General-Purpose Large Language Models Generalize to English-Thai Machine Translation?}

\author{
 \textbf{Jirat Chiaranaipanich\textsuperscript{1}},
 \textbf{Naiyarat Hanmatheekuna\textsuperscript{2}},
 \textbf{Jitkapat Sawatphol\textsuperscript{3}},
 \textbf{Krittamate Tiankanon\textsuperscript{4}},
\\
 \textbf{Jiramet Kinchagawat\textsuperscript{4}},
 \textbf{Amrest Chinkamol\textsuperscript{3,4}},
 \textbf{Parinthapat Pengpun\textsuperscript{5}},
 \textbf{Piyalitt Ittichaiwong\textsuperscript{4,6,7,*}},
\\
 \textbf{Peerat Limkonchotiwat\textsuperscript{3,*}},
\\
\\
 \textsuperscript{1}Ruamrudee International School, 
 \textsuperscript{2}Chulalongkorn University,
 \textsuperscript{3}Vidyasirimedhi Institute of Science and Technology,\\
 \textsuperscript{4}PreceptorAI team, CARIVA Thailand, 
 \textsuperscript{5}Bangkok Christian International School,
 \textsuperscript{6}Mahidol University, \\
 \textsuperscript{7}King's College London,
 \textsuperscript{*}Corresponding authors\\
 \texttt{piyalitt.itt@preceptorai.tech},
\texttt{peerat.l\_s19@vistec.ac.th}
}

\begin{document}
\maketitle
\begin{abstract}
Large language models (LLMs) perform well on common tasks but struggle with generalization in low-resource and low-computation settings. We examine this limitation by testing various LLMs and specialized translation models on English-Thai machine translation and code-switching datasets. Our findings reveal that under more strict computational constraints, such as 4-bit quantization, LLMs fail to translate effectively. In contrast, specialized models, with comparable or lower computational requirements, consistently outperform LLMs. This underscores the importance of specialized models for maintaining performance under resource constraints.

\end{abstract}

\section{Introduction}

Large language models (LLMs) have shown remarkable capabilities in Neural Machine Translation (NMT) and code-switching (CS), attributed to their robustness and generalization \citep{vaswani2013decoding,naveed2024comprehensiveoverviewlargelanguage,Radford2019LanguageMA}. Recent studies indicate that NMT and CS are largely solved for LLMs in high-resource languages \citep{DBLP:journals/corr/abs-2004-05809,Hamed2017-ja,mandarincodeswitch}. However, our research reveals that this performance fails to generalize to low-resource and low-computation settings, which is critical for real-world settings where computational resources are constrained.

This paper explores the generalization of LLMs through two research questions: (i) How do general-purpose LLMs and specialized translation models generalize to low-resource language translation? (ii) How do real-life computational constraints affect performance metrics? To address these questions, we experiment with Llama-3 in various quantization settings. Additionally, we compare LLMs with specialized translation models like NLLB \cite{nllbteam2022languageleftbehindscaling} to evaluate performance and efficiency trade-offs.

\section{Experimental Setup}
\textbf{Datasets}. We evaluated two translation datasets: (i) a proprietary medical CS translation dataset\footnote{\url{https://cariva.co.th/}}, containing 63,982 English-Thai sentence pairs with retained English medical terms; and (ii) scb-mt-en-th-2020 \citep{Lowphansirikul_2021}, a 1,001,752 sentence pair English-Thai translation dataset, from which we randomly selected 63,982 pairs to match the sample size of the CS dataset.

\noindent \textbf{Models} Our evaluation focused on three models pertinent to our research questions: Llama-3 8B \citep{dubey2024llama3herdmodels}, NLLB-600M, and NLLB-3.3B \citep{nllbteam2022languageleftbehindscaling}. For the Llama-3 model, we assessed both the pre-trained and finetuned versions, with the latter quantized to 2, 3, 4, and 8 bits using GPTQ~\cite{gptq}. For the NLLB models, we evaluated both pre-trained and finetuned versions. All were finetuned for 3 epochs with a learning rate of 2e-4 on an A100 GPU.

\noindent \textbf{Metrics} We employed standard MT metrics for evaluation, such as BLEU3, METEOR, and CER. Additionally, we measured the CS boundary F1 score, which is the harmonic mean of precision and recall for correctly preserved English terms \citep{sterner-teufel-2023-tongueswitcher}.

\noindent \textbf{LLM-as-a-judge Evaluation}. To analyze performance degradation, we used GPT4-o\footnote{snapshot gpt-4o-2024-05-13} as a judge with 3-shot prompting to identify failure modes in each predicted translation. GPT4-o received the source, target, and predicted sentences. The LLM judge assigned a multiple-choice label to each translation, categorizing them as "Forgot to translate," "Meaning changed," "Gibberish," or "Excellent," with a "Keywords not preserved" category for the CS translation task.

\section{Results}

As illustrated in Table \ref{tab:evaluation_results_combined}, NLLB-3.3B and NLLB-600M outperform Llama-3 8B on most metrics, despite using 2.35x and 10.81x less VRAM, respectively. 
This contrasts with prior studies indicating the superiority of general-purpose language models in specialized, low-resource tasks \citep{li2023chatgptgpt4generalpurposesolvers,nori2023generalistfoundationmodelsoutcompete,naveed2024comprehensiveoverviewlargelanguage}.
%
Moreover, the average percentage difference between NLLB-3.3B and full-precision Llama-3 8B across BLEU and METEOR scores is $\sim$23.39\% and $\sim$1.33\% for the SCB and CS dataset, respectively. This minimal difference for the CS dataset suggests that NLLB's multilingual pre-training is not a significant advantage in translation-adjacent tasks.

Interestingly, Llama-3-8B excels in the METEOR metric for CS translation, which accounts for word stems and synonyms. This suggests Llama-3-8B produces relevant but imprecise translations, affecting metrics that require exact matches but not METEOR.

\newcolumntype{P}[1]{>{\centering\arraybackslash}m{#1}}
\begin{table}[h!]
    \centering
    \vspace{-2mm}
    \begin{scriptsize}
    \begin{tabular}{P{7mm}|c|P{4.5mm}P{6.5mm}P{3.5mm}P{3mm}P{5.5mm}}
    \hline
    Dataset & Model Variant & BLEU3 & METEOR & CER & CS-F1 & Memory (GB) \\ \hline
    \multirow{7}{*}{CS} & Llama-3-8b & 0.421 & 0.615 & 6.606 & 0.330 & 31.48 \\
    & Llama-3-8b-8bit & 0.421 & \textbf{0.616} & 6.622 & 0.332 & 9.87\\
    & Llama-3-8b-4bit & 0.392 & 0.591 & 6.833 & 0.320 & 7.13 \\
    & Llama-3-8b-3bit & 0.214 & 0.410 & 8.437 & 0.280  & 5.42 \\
    & Llama-3-8b-2bit & 0.001 & 0.013 & 4.565 & 0.002 & 4.48\\ 
    & NLLB-3.3b & \textbf{0.443} & 0.600 & \textbf{0.419} & \textbf{0.398} & 13.42\\
    & NLLB-0.6b & 0.410 & 0.576 & 0.438 & 0.394 & 2.91\\ \hline
    \multirow{7}{*}{SCB}
    & Llama-3-8b & 0.173 & 0.371 & 30.416 & - & 31.48 \\
    & Llama-3-8b-8bit & 0.173 & 0.371 & 30.952 & - & 9.46 \\
    & Llama-3-8b-4bit & 0.156 & 0.349 & 30.576 & - & 7.14\\
    & Llama-3-8b-3bit & 0.079 & 0.231 & 31.088 & - & 5.35\\
    & Llama-3-8b-2bit & 0.000 & 0.003 & 19.232 & - & 4.44\\
    & NLLB-3.3b & \textbf{0.244} & \textbf{0.449} & 0.585 & - & 13.31 \\
    & NLLB-0.6b & 0.238 & 0.437 & \textbf{0.574} & - & 2.91 \\  \hline
    \end{tabular}
    \vspace{-2mm}
    \caption{Evaluation Results for LLMs and specialized translation models on CS and SCB datasets. }
    \vspace{-2mm}
    \label{tab:evaluation_results_combined}
    \end{scriptsize}
    \vspace{-3mm}
\end{table}

\begin{figure}[!htp]
        
    \centering
    \begin{subfigure}[b]{0.4\textwidth} 
        \centering
        \caption{CS LLM Judge Grading}
        \includegraphics[width=\textwidth]{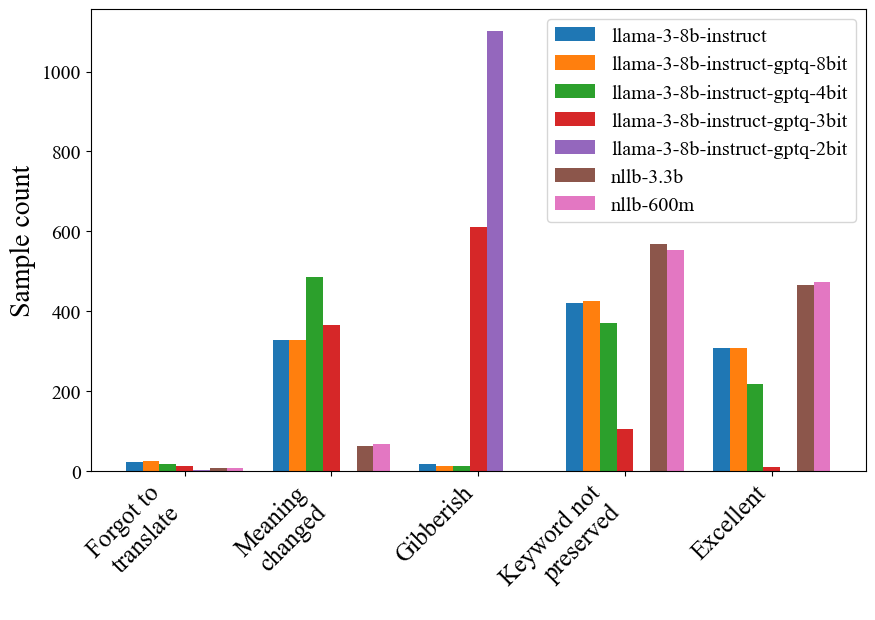}
        \vspace{-5mm}
        \label{fig:csllmjudgecomparison}
    \end{subfigure}
    \hspace{1em} 
    
    \begin{subfigure}[b]{0.4\textwidth} 
        \centering
        \caption{SCB LLM Judge Grading}
        \includegraphics[width=\textwidth]{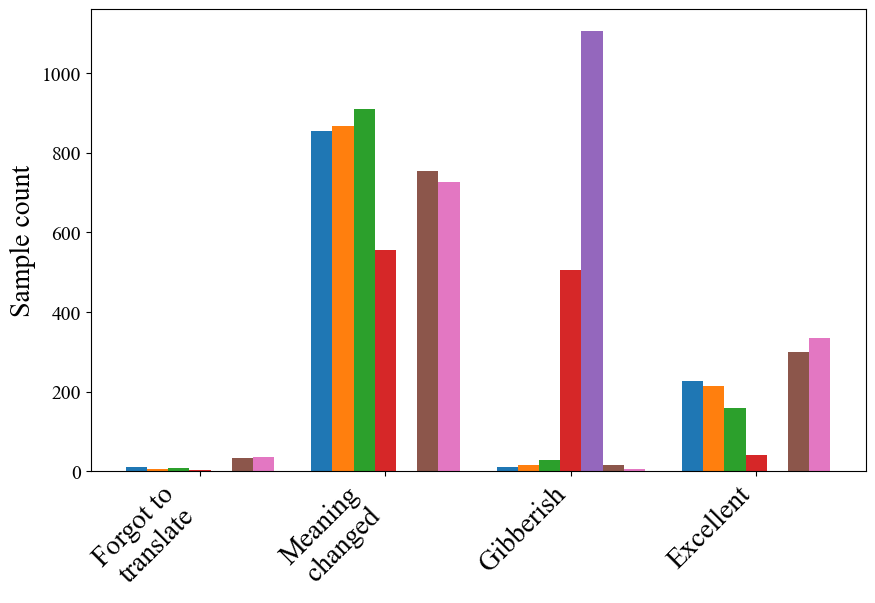}
    \label{fig:scbllmjudgecomparison}
    \end{subfigure}
    \vspace{-7mm}
    \caption{Llama-3 and NLLB failure analysis. Note that the legend is shared between Figures~\ref{fig:csllmjudgecomparison} and~\ref{fig:scbllmjudgecomparison}.}
    \label{fig:combined}
    \vspace{-8mm}
\end{figure}

\section{Analysis}
\noindent 
\textbf{Failure Analysis} As shown in Figures~\ref{fig:csllmjudgecomparison} and ~\ref{fig:scbllmjudgecomparison}, we observed a divergence in failure modes between the two datasets. For the SCB dataset, errors initially rise in the "Meaning changed" category (from 16 to 4 bits) and then in the "Gibberish" category (from 4 to 2 bits). In the CS dataset, errors first increase in the "Meaning changed" category while decreasing in "Keywords not preserved" (from 16 to 4 bits), followed by an increase in "Gibberish" errors (from 4 to 2 bits). Notably, the best-performing models (NLLB-3.3B and NLLB-600M) exhibit the highest number of "Forgetting to preserve" errors. This suggests an alternative failure mode in CS translation, where top models first lose the ability to preserve medical keywords, then to translate accurately, and finally to translate at all. Importantly, despite higher errors in the "Forgetting to preserve" category, NLLB models perform better on the CS-F1 metric, Table~\ref{tab:evaluation_results_combined}, highlighting the importance of task-specific metrics. 

\noindent \textbf{Impact of Quantization}
Interestingly, CS results show greater resilience to quantization than SCB results. Across BLEU, CER, and METEOR metrics, CS translation results experience less degradation than SCB results when compared against the full-precision baseline. This may be due to the early loss of complex Thai vocabulary during quantization, while complex English vocabulary, rewarded in the CS task, is better preserved. The resilience of CS results suggests a novel approach for mitigating performance degradation in quantized multilingual models by leveraging CS outputs.

\section{Conclusion}
We study the performance of general-purpose and specialized language models on translation and translation-adjacent tasks. Our findings indicate that specialized translation models outperform general-purpose models, although the performance gap is smaller for CS translation. As models undergo increased quantization, the divergence in failure modes between SCB and CS datasets underscores the importance of task-specific metrics.

\bibliography{custom}


\section{Appendix}
\label{sec:appendix}

\begin{table*}[!ht]
    \centering
    \caption{Full Evaluation Result on the CS and SCB datasetes. ''Memory(GB)'' indicates the memory consumption for single-batch inference on an A100 GPU. ''Runtime vs 16bit Llama'' represents the inference time speedup compared to a 16bit Llama baseline.}
    \vspace{-2mm}
    \label{tab:eval_result} %
    \begin{scriptsize}
   \begin{tabular}{l|l|llllllll}
    \hline
Dataset & Model Variant & BLEU3 & METEOR & CER & WER & chrF & CS-F1 & Memory (GB)  & Runtime vs 16bit Llama (\%)\\ \hline
    \multirow{7}{*}{CS}
    & Llama-3-8b & 0.421 & 0.615 & 6.606 & 0.526 & 0.402 & 0.330 & 31.48  &0\\
    & Llama-3-8b-8bit & 0.421 & \textbf{0.616} & 6.622 & 0.525 & 0.401 & 0.332 & 9.87  &-17.08\\
    & Llama-3-8b-4bit & 0.392 & 0.591 & 6.833 & 0.559 & 0.386 & 0.320 & 7.13  &-61.05\\
    & Llama-3-8b-3bit & 0.214 & 0.410 & 8.437 & 0.917 & 0.262 & 0.280 & 5.42  &30.11\\
    & Llama-3-8b-2bit & 0.001 & 0.013 & 4.565 & 4.616 & 0.039 & 0.002 & 4.48  &-15.17\\ 
    & NLLB-3.3b & \textbf{0.443} & 0.600& \textbf{0.419} & \textbf{0.460}& \textbf{0.571} & \textbf{0.398}& 13.42  &-85.66\\
    & NLLB-0.6b & 0.410 & 0.576 & 0.438 & 0.487 & 0.551 & 0.394 & \textbf{2.91}&\textbf{-97.13}\\ \hline
    \multirow{7}{*}{SCB}
    
    & Llama-3-8b & 0.173 & 0.371 & 30.416 & 0.865 & 0.147 & - & 31.48  &0
\\
    & Llama-3-8b-8bit & 0.173 & 0.371 & 30.952 & 0.867 & 0.145 & - & 9.46  &23.82
\\
    & Llama-3-8b-4bit & 0.156 & 0.349 & 30.576 & 0.891 & 0.138 & - & 7.14  &-55.41
\\
    & Llama-3-8b-3bit & 0.079 & 0.231 & 31.088 & 1.142 & 0.105 & - & 5.35  &-22.83
\\
    & Llama-3-8b-2bit & 0.000 & 0.003 & 19.232 & 20.030 & 0.004 & - & 4.44  &22.44\\ 
    & NLLB-3.3b & \textbf{0.244} & \textbf{0.450} & \textbf{0.585} & 0.729 & \textbf{0.475} & - & 13.31  &-86.52
\\
    & NLLB-0.6b & 0.238 & 0.437 & 0.574 & \textbf{0.721}& 0.461 & - & \textbf{2.91}&-\textbf{97.05}\\ \hline
    \end{tabular}
    \end{scriptsize}
    \vspace{-3mm}
\end{table*}
\subsection{Finetuning Prompts for Llama}
\textbf{Code-switching (CS) Prompt}

\begin{lstlisting}
You are a helpful code switching English to Thai language translation assistant. Translate the given English texts to Thai while preserving the medical keywords.
\end{lstlisting}
\textbf{Machine translation (SCB) Prompt}
\begin{lstlisting}
You are a helpful English to Thai language translation assistant. Translate the given English texts to Thai.
\end{lstlisting}
\subsection{LLM Judge Prompts}
\textbf{LLM Judge Code-switching Dataset Prompt}
\begin{lstlisting}
You will be given a user_text, model_answer, and system_translation trio. Your task is to provide a multiple choice answer, analyzing the cause of failure of the system's translation of the user's text when compared to the model_answer.Give your answer letter which can either be A, B, C, D, E.

Here are the choices.
A: The system_translation forgot to translate: missed translating a large part of the text
B: The system_translation translated wrongly: adds additional information or hallucinates; changes the meaning in some significant way
C: The system_translation is gibberish: it does not make sense and is just a jumble of words and characters
D: The system_translation forgot to preserve the CS keyword: although the text is translated; the meaning is quite well preserved; the keywords are translated amd not preserved in the orignal language
E: The system_translation is excellent: preserves the keywords; has almost the meaning as the model answer; everything except for the keywords are translated

You MUST provide the answer letter. Do not provide anything else.

Here are examples with the best answer given plus reasoning.

EXAMPLE 1:
User Text: USER_TEXT_1
Model Answer: MODEL_ANSWER_1
System Translation: SYSTEM_TRANSLATION_1
Reasoning: REASONING_1

EXAMPLE 2:
User Text: USER_TEXT_2
Model Answer: MODEL_ANSWER_2
System Translation: SYSTEM_TRANSLATION_2
Reasoning: REASONING_2

EXAMPLE 3:
User Text: USER_TEXT_3
Model Answer: MODEL_ANSWER_3
System Translation: SYSTEM_TRANSLATION_3
Reasoning: REASONING_3

Below are the text, answer, and translation. Give a multiple choice response.

User Text: {user_text}
Model Answer: {model_answer}
System Translation: {system_translation} 
\end{lstlisting}

\textbf{LLM Judge Machine Translation Dataset (SCB) Prompt}
\begin{lstlisting}
You will be given a user_text, model_answer, and system_translation trio.
Your task is to provide a multiple choice answer, analyzing the cause of failure of the system's translation of the user's text when compared to the model_answer.
Give your answer letter which can either be A, B, C, D.

Here are the multiple choices.
A: The system_translation forgot to translate: missed translating a large  part of the text
B: The system_translation translated wrongly: adds additional information or hallucinates; changes the meaning in some significant way
C: The system_translation is gibberish: it does not make sense and is just a jumble of words and characters
D: The system_translation is excellent: has almost the meaning as the model answer, everything is translated

You MUST provide the answer letter. Do not provide anything else other than the multiple choice answer letter.

Here are a few examples with the best multiple choice answer given plus reasoning.

EXAMPLE 1:
User Text: USER_TEXT_1
Model Answer: MODEL_ANSWER_1
System Translation: SYSTEM_TRANSLATION_1
Reasoning: REASONING_1

EXAMPLE 2:
User Text: USER_TEXT_2
Model Answer: MODEL_ANSWER_2
System Translation: SYSTEM_TRANSLATION_2
Reasoning: REASONING_2

EXAMPLE 3:
User Text: USER_TEXT_3
Model Answer: MODEL_ANSWER_3
System Translation: SYSTEM_TRANSLATION_3
Reasoning: REASONING_3

EXAMPLE 4:
User Text: USER_TEXT_4
Model Answer: MODEL_ANSWER_4
System Translation: SYSTEM_TRANSLATION_4
Reasoning: REASONING_4

Below are the user text and system translation pair. Give a multiple choice response.

User Text: {user_text}
Model Answer: {model_answer}
System Translation: {system_translation}
\end{lstlisting}

\end{document}